\documentclass[twocolumn,conference,a4paper]{IEEEtran}
\usepackage{amssymb,amsmath,epsfig,latexsym,graphicx,bm,cite}
\usepackage[flushleft]{threeparttable}
\usepackage[left = .56in, top=.75in,right=.56in,bottom=0.8in,nohead,nofoot]{geometry}
\usepackage{multirow}
\usepackage{subfloat}
\usepackage{epstopdf}

\begin{document}

\title{
	Detection of Inferior Myocardial Infarction using Shallow Convolutional Neural Networks
}
\author{\IEEEauthorblockN{Tahsin Reasat}
	\IEEEauthorblockA{Department of Electrical and Electronic Engineering\\
		Bangladesh University of Engineering and Technology\\
		Dhaka, Bangladesh\\
		Email: tahsinreasat@ieee.org, greasat@gmail.com}
	\and
	\IEEEauthorblockN{Celia Shahnaz}
	\IEEEauthorblockA{Department of Electrical and Electronic Engineering\\
		Bangladesh University of Engineering and Technology\\
		Dhaka, Bangladesh\\
		Email: celia@eee.buet.ac.bd}
	}

\maketitle
\begin{abstract}
Myocardial Infarction is one of the leading causes of death worldwide. This paper presents a Convolutional Neural Network (CNN) architecture which takes raw Electrocardiography (ECG) signal from lead II, III and AVF and differentiates between inferior myocardial infarction (IMI) and healthy signals. The performance of the model is evaluated on IMI and healthy signals obtained from  Physikalisch-Technische Bundesanstalt (PTB) database. A subject-oriented approach is taken to comprehend the generalization capability of the model and compared with the current state of the art. In a subject-oriented approach, the network is tested on one patient and trained on rest of the patients. Our model achieved a superior metrics scores (accuracy= 84.54\%, sensitivity= 85.33\% and specificity= 84.09\%) when compared to the benchmark. We also analyzed the discriminating strength of the features extracted by the convolutional layers by means of geometric separability index and euclidean distance and compared it with the benchmark model.
\end{abstract}

\section{INTRODUCTION}
Myocardial Infarction (MI), commonly known as `heart attack', is the death of heart muscles due to prolonged lack of oxygen supply (ischemia). Myocardial infarction is the leading cause of death in the United States and in most industrialized nations throughout the world. In 2014, on average, someone in USA died  every 4 minutes due to  a stroke \cite{aha}. Survival rates improve after a heart attack if treatment begins within 1 hour. This signifies the necessity of accurate and timely diagnosis of MI. A myocardial infarction is clinically defined as \cite{mi_criteria}:
	\begin{itemize}
		\item Elevated blood levels of cardiac enzymes (CKMB or Troponin T)
	\end{itemize}
	Additionally, one of the following criteria have to be met:
	\begin{itemize}
		\item The patient has typical complaints,
		\item The ECG shows ST elevation or depression.
		\item pathological Q waves develop on the ECG
		\item A coronary intervention had been performed (such as stent placement)
	\end{itemize}
Although detection of elevated serum cardiac enzymes is more important than ECG changes, the cardiac enzymes can only be detected in the serum 5-7 hours after the onset of the myocardial infarction. Therefore, in the first few hours after the myocardial infarction, the ECG can be crucial. MI can be diagnosed by cardiologists based on the changes in the ECG, but the sensitivity and specificity of manual detection of acute MI is  91\% and 51\% as reported in \cite{mi_manual}. Developing a computer aided system to automatically detect MI would help the cardiologists make better decisions. Hence, various researches has been conducted on MI detection. These researches has focused on wavelet transform-based methods \cite{dwt_1,sharma} time-domain-based algorithm \cite{td_1}, ST-segment-based technique \cite{st_3}, hidden Markov model-based technique \cite{markov} and Vectorcardiogram based method \cite{vector_1}. 
	
These works have taken a hand-crafted feature extraction approach for the classification problem. It would be convenient if the process of feature extraction/selection could also be automated. With this intention in mind, researchers have tried to borrow elements from \textit{deep learning} and apply them to ECG signals. The key advantage of deep learning is that the model itself learns the most discriminative features from raw data and tries to match its output with the desired result. Deep learning algorithms, at its inception, were mainly applied in computer vision \cite{lecun,alexnet}. In recent years, its application has quickly spread to many sectors including ECG signal. Convolutional neural networks (CNN) have been utilized in arrhythmia detection, coronary artery disease detection, beats classification \cite{cnn_arrythmia,cnn_cad,cnn_beat}. Deep belief network has been used to classify signal quality in ECG \cite{dbn_signal_quality}. Recurrent neural networks (RNN) have also been used in beats classification, obstruction of sleep apnea detection, ecg-based biometrics \cite{rnn_beats,rnn_apnea,rnn_biometrics} 
	
In \cite{acharya} researchers have implemented an 11 layer CNN to detect MI. Although the model achieved satisfactory results, it has been shown that deep learners are not necessary in ECG signal analysis \cite{cnn_beat}. By downsampling the ECG signal, the researchers were able to train a shallow CNN and achieve superior results in beat classification. Hence, there is scope of improvement regarding the reduction of network size and training complexity. Additionally, the experiment in \cite{acharya} was based on a \textit{class-oriented} approach, i.e., heartbeats extracted from ECG signals were randomly sampled to form test and train set. As a result samples from the same patient contained both in test and train. Therefore, it is not clear how the model will perform on patients it has never seen before. Hence, to test the model's generalization capability a \textit{subject-oriented} approach \cite{sharma} should be taken. In this approach, the heartbeats are grouped according to the patients. The model is tested on heartbeats from one patient while it is trained on heartbeats from rest of the patients. This procedure is repeated for the remaining the subjects.
Myocardial infarction can be classified according to the location of the infarct. This paper focuses on the detection of inferior myocardial infarction(IMI). We implement a shallow but wide CNN whose architecture resembles the inception module in GoogleNet \cite{inception}. A subject-oriented approach is taken to evaluate the model's performance on the Physikalisch-Technische Bundesanstalt (PTB) dataset \cite{ptbdb} collected from PhysioNet \cite{physionet}. It's performance is compared with the current state of the art \cite{sharma} where IMI detection is done using features extracted from the stationary wavelet transform (SWT) of the ECG signal.  We also compare the measure of class separability of the two methods in terms of geometric separability index and euclidean distance. The paper is organized in the following way: In section \ref{method}  the proposed method is described, in section \ref{exp_result} the experimental results are discussed, in section \ref{qf} quality of feature is compared between the proposed and benchmark model and the conclusion is drawn in section \ref{conclusion}. 
	
\section{METHODOLOGY}\label{method}
\subsection{Data}
The database contains 549 records from 290 subjects. Out of 290 subjects there are 148 MI subjects and 52 healthy controls (HC). Thirty of the 148 MI patients have IMI; they  are included in the experiment along with the healthy controls. Each subject is represented by one to five records. Each record includes 15 simultaneously measured signals: the conventional 12 leads and three frank leads. Each signal is digitized at 1000 Hz. The change in ECG signal due to IMI is more prominent in Lead II, III and AVF \cite{mi_lead}. Hence, these three leads from each ECG record was used in this work.
	
\subsection{Data Processing}
Each signal were downsampled from 1kHz to 250Hz. A two stage median filter was used to remove baseline wandering. The local window size for the first and second stage was chosen to be 125 and 249 respectively. Next, Savitzky-Golay (SG) smoothing filter with order 3 and frame size 15 was used to remove noise. Up to this point the preprocessing steps are identical to \cite{sharma}. The denoised signal was further downsampled to 64Hz to decrease computational burden on the convolutional network and speed up training time. Finally, the signals were partitioned into short segments of 3.072 seconds (196 signal samples per segment). The segments generated simultaneously from three different leads were grouped together to create a sample. This yielded a total of 3222 IMI samples and 3055 HC samples. 

\subsection{Network Architecture}
\begin{figure}
	\centering
	\includegraphics[width=3in]{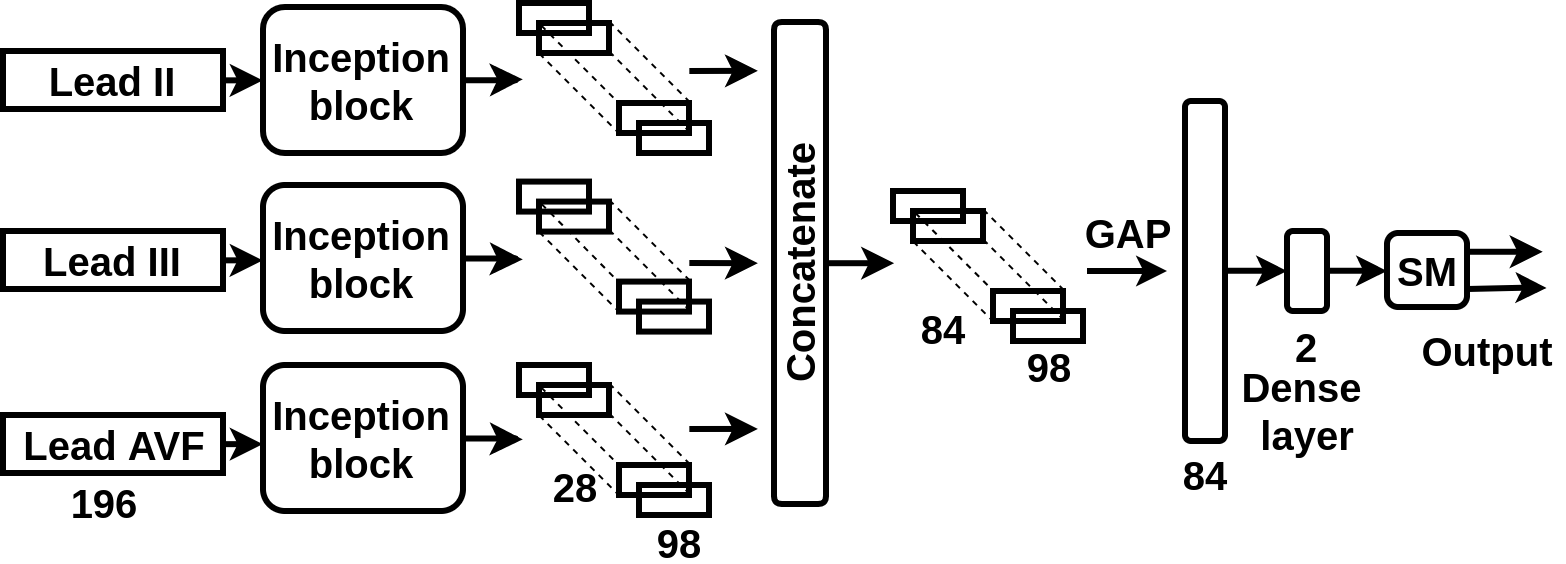}\\
	\caption{The model architecture. Here GAP stands for Global Average Pooling and SM stands for Softmax.}\label{fig_model}
\end{figure} 

\begin{figure}
	\centering
	\includegraphics[width=3in]{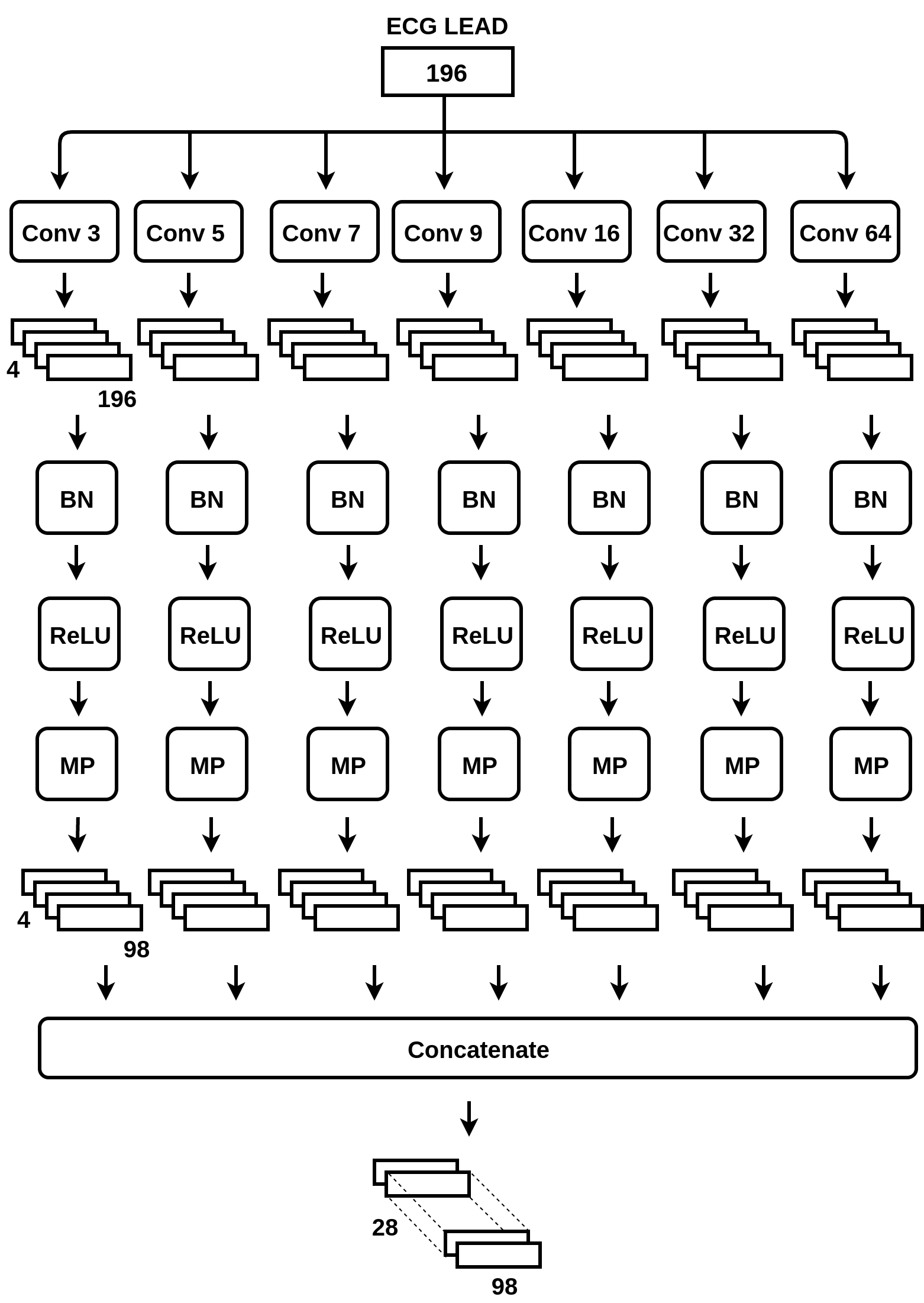}\\
	\caption{The inception block. Each `Conv $n$' layer  has 4 filters with window length $n$, ($n\in\{3,5,7,9,16,32,64\}$). BN and MP stands for Batch Normalization and Max Pooling, respectively.}\label{fig_inception}
\end{figure} 

The high-level architecture of the model is illustrated in fig. \ref{fig_model}. The network takes raw ECG sample at the input layer. Each sample consists of signal segments from three leads, lead II, III and AVF. Each lead is fed into an \textit{inception block} (fig. \ref{fig_inception}). In the inception block the input goes through seven parallel paths. In each parallel path there is a convolutional layer followed by a batch normalization layer, a rectified linear unit (ReLU) activation layer and a max pooling layer. Feature maps extracted by the inception blocks are concatenated and passed on to a global average pooling layer. Finally, there is a 2 unit dense layer with a softmax activation layer which gives the categorical probability. The weight of the dense layer is L2 regularized to prevent it from overfitting. The motivation behind the structure of the key layers are described below. 

\subsubsection{Convolutional Layer}
The function of the convolutional layer is to extract features from the input. In traditional CNNs, the filters of a particular layer has the same window length which is gradually reduced in the subsequent layers \cite{alexnet}. However, there are no hard and fast rules to select the window length and it is selected experimentally. Researchers in \cite{inception} have used varying window size in the same layer allowing it to look at image patches of different size. So the network itself learns the features from different filter size while training on the data. Inspired by this architecture we have used different window length for each of the convolution layers in the seven parallel paths. There are 4 filters in each path and the filter size vary from 3 to 64 as shown in fig \ref{fig_inception}. This enables the inception block to look at the ECG signal at multiple spatial resolution and extract multi-level features from the same input. The convolutional filters are gradually shifted by one sample and the input signal is zero padded to keep the output length unaltered (196 samples).

\subsubsection{Batch Normalization Layer}
The distribution of the output of the convolutional layer changes as the training parameters changes. The following layers have to continuously adapt to new distributions which requires it to have a slow learning rate. This phenomenon is referred to as \textit{internal covariate shift} \cite{batchnorm}. To tackle this issue a batch normalization layer is included after the convolutional layer. This layer includes normalization step that fixes the means and variances of the following layer inputs. As a result, higher learning rate can be used resulting in faster training time. Batch normalization layer also works as a regularizer \cite{batchnorm}.
\subsubsection{ReLU layer}
The rectified linear \cite{relu} layer induces a nonlinearity in the values of the incoming layer. The function of ReLU can be summarized mathematically as $f(x)=max(x,0)$ i.e. it only passes the values $x$ which are greater than zero.
\subsubsection{Max Pooling Layer}
The max pooling layer downsamples the input. This reduces computational cost due the decrease in dimension and provides translational invariance to the internal representation. In our model both the size of the pooling window and its stride was chosen to be two samples.
\subsubsection{Global Average Pooling Layer}\label{gap}
In traditional CNNs, convolutional layers are followed by dense layers which transforms the feature maps into the desired output. However, the dense layers act as black-boxes which make it difficult to interpret the connection between filters and categorical outputs. Additionally, dense layers are prone to overfitting and computationally expensive due to its large number of parameters \cite{globalaveragepooling}. Hence, global average pooling layer is used which calculates the spatial average of each feature map. It summarizes the feature map extracted by each filter into a single value without the need of learning any parameters. As a result, we get 84 features from 84 filters which are passed on to the dense layer for classification.

\subsection{Metrics}
The performance of the model was evaluated in terms of accuracy(Ac), sensitivity(Se), and specificity(Sp) of the test predictions. 
The metrics are defined as
	\begin{align}
		Ac\%= \frac{tp+tn}{tp+tn+fp+fn}*100\\
		Se\%= \frac{tp}{tp+fn}*100\\
		Sp\%= \frac{tn}{tn+fp}*100
	\end{align}
Here, $tp$ is true positive prediction, $fp$ is false positive prediction, $tn$ is true negative prediction and $fp$ is false negative prediction.
\subsection{Train-Test Split}
Splitting of the train and test set is performed in a subject-oriented approach. In this approach the segments are grouped according to patients. Testing is done on one patient and training is done on the remaining 81 patients. The same procedure was followed for the other patients. Average accuracy, sensitivity and specificity is reported in this paper.
\section{Experimental Results}\label{exp_result}
The network was implemented using Keras \cite{keras} neural network library. The weights of the dense hidden layer was regularized using l2 regularization with regularization parameter $\lambda=0.001$ to prevent the model from overfitting. Categorical crossentropy was used as the loss function. For a 2 class problem the crossentropy is given by  
\begin{align}
 \mathcal{L}(p,y)=-ylog(p)-(1-y)log(1-p)
\end{align}
Here, $y\in\{0,1\}$ represents the true labels of each sample and $p$ is the probabilty given by the model. The model was trained using backpropagation algorithm and the adam \cite{adam} optimizer was used to update the weights. The network was trained with an initial learning rate of 1e-3 and varied in the range [1e-3,1e-5]. The exponential decay rate for the first moment estimates $\beta_{1}$ and the exponential decay rate for the second-moment estimates $\beta_{2}$ was chosen to be 0.9 and 0.999, respectively. The learning rate was scheduled to be decreased by a factor of 10 if there were no improvements in training loss for 5 consecutive epochs. Training was stopped if there was no improvement in training loss for 10 consecutive epochs. The training was allowed to run for a maximum of 200 epochs. The network was trained in mini-batches with batch size 32. The scores of the proposed CNN and the benchmark method is summarized in table \ref{table_subject}.
	\begin{table}[h]
		\begin{center}
				\caption{Comparison of the proposed technique with existing method in subject-oriented approach}
				\label{table_subject}
				\begin{tabular}{|c|c|c|c|}
					\hline
					Method & Avg. Ac\% & Se\% & Sp\%\\
					\hline
					SWT + KNN\cite{sharma}  &75.80 &75.62 &73.66 \\
					\hline
					SWT + SVM\cite{sharma} &81.71 &79.01 &79.26\\
					\hline
					CNN on raw ECG & 84.54 & 85.33 & 84.09\\
					\hline
				\end{tabular}
			
		\end{center}
	\end{table}

\begin{table}[h]
	\begin{center}	
		\caption{confusion matrix for the cnn's prediction when trained on all the patients}
		\label{table_cf}
		\begin{tabular}{l|l|c|c|}
			\multicolumn{2}{c}{}&\multicolumn{2}{c}{Predicted Class}\\
			\cline{3-4}
			\multicolumn{2}{c|}{}&HC&IMI \\
			\cline{2-4}
			\multirow{2}{*}{Actual Class}& HC & $2833$ & $222$\\
			\cline{2-4}
			& IMI & $10$ & $3212$ \\
			\cline{2-4}
			
\end{tabular}
	\end{center}
	\end{table}
\section{Quality of Features} \label{qf}
As mentioned before in subsection \ref{gap}, the global average pooling layer compresses each incoming filter map into a single feature. Therefore, output of the global average pooling layer can be considered as the final feature vector upon which the dense output layer makes its classification decision. We use Thornton's geometric separability index (GSI)\cite{gsi} to measure the degree to which samples associated with the same output cluster together. It is given by
\begin{align}
GSI(f)=\dfrac{\sum_{i=1}^{N} f(\mathbf{x}_{i})+f(\mathbf{x}_{i}^{'})+1~mod~2}{N}
\end{align}
Here, $\mathbf{x}_{i}=\{x_i^1,x_i^2,...,x_i^L\}$ is the feature vector with dimension $L$, $\mathbf{x}_{i}^{'}$ is the nearest neighbor of $\mathbf{x}_{i}$, $f$ is a binary target function, and $N$ is the total number of samples. In this work, the nearest neighbor function utilizes Euclidean distance between a pair of feature vectors.

To find out the similarity between features of the same class, we calculate the average of the euclidean distance ($D_{E}^{k}$) between each pair of samples within the same class.
\begin{align}
D_{E}^{k}=\dfrac{(|\mathcal{N}_k|-2)!2!}{|\mathcal{N}_k|!}\sum_{\substack{\mathbf{x}_i,\mathbf{x}_j\in \mathcal{N}_k\\i \neq j}}\sqrt{\sum_{l=1}^{L}(x_{i}^{l}-x_{j}^{l})^2}
\end{align}
Here, $\mathcal{N}_k$ is the set of feature vectors belonging to class $k$, $k\in\{HC,IMI\}$. $L$ is the dimension of the feature vector. To calculate GSI and $D_E^k$ of the benchmark model we implement the stationary wavelet transform based model as described in \cite{sharma} and compute the features for each sample. We also trained our network on all of the available samples following the procedure explained in section \ref {exp_result}. The confusion matrix for the prediction performance of the trained model is shown in table \ref{table_cf}. For each of the samples, output of the global average pooling layer were extracted and treated as the feature vector. 

The GSI and $D_E^k$ of the proposed and benchmark model is compared in table \ref{table_separability}. The GSI is high for both models, signifying that the feature vectors of the same class are located in a close cluster. The GSI of the proposed model is slightly larger than the GSI of the benchmark. For the healthy samples, the $D_E^{HC}$ of the proposed model is slightly higher than the benchmark model. For the IMI samples, the $D_E^{IMI}$ of the proposed model is much lower than the benchmark model. Overall, the features extracted from proposed model shows good discriminating strength. Hence, the proposed CNN does a better job in differentiating between the two classes.

\begin{table}
	\centering
	\caption{Comparison of $D_E^k$ between benchmark and proposed model}\label{table_separability}
	\begin{tabular}{|c|c|c|c|}
		\hline
		Method & GSI & $D_{E}^{HC}$& $D_{E}^{IMI}$\\
		\hline
		SWT & 0.9852 & 3.10 & 3.71\\
		\hline
		CNN & 0.9987 & 3.30 & 2.69\\
		\hline
	\end{tabular}
\end{table}
  
\section{CONCLUSIONS} \label{conclusion}
	
In this paper we have demonstrated the use of a shallow convolutional neural network in the detection of inferior myocardial infarction. This network benefits from the use of varying filter size in the same convolution layer which allows it to learn features from signal regions of varying length. Future research should focus on the effect of varying the filter length as well as increasing the number of inception layers. Another important research direction is the study of the relationship between the extracted features and the actual ECG segments. Finding out which portions of the ECG signal activates the filters would lead to a better understanding of the disease itself. And lastly this research only focuses on the detection of inferior myocardial infarction. Classification of different infarctions based on their positions should be investigated in future works.

\bibliographystyle{IEEEtran}
\bibliography{CNN_IMI}
		
\end{document}